\documentclass{IETEJR}

\usepackage{amsmath,amsfonts,amssymb}
\usepackage{graphicx}
\usepackage{setspace}
\usepackage{mathtools}
\usepackage{lineno,hyperref}
\usepackage{kantlipsum}
\usepackage{epsfig} 
\usepackage{tocloft}
\usepackage{enumerate}
\usepackage{booktabs}
{}
{}
{}
\usepackage[authoryear]{natbib}

\begin{document}

\title{Feature Based Framework to Detect Diseases, Tumor, and Bleeding in Wireless Capsule Endoscopy}

\author{Omid Haji Maghsoudi$^{1*}$, and Mahdi Alizadeh$^{2}$
\vspace*{0.05em}\\\small $^{1}$Department of Bioengineering, \\
\small Temple University, 12th Street, Philadelphia, PA, USA \\
\small $^{*}$o.maghsoudi@temple.edu \\
\vspace*{0.05em} \small $^{2}$Stereotactic neuroimaging specialist, \\
\small Thomas Jefferson Hospital, Philadephia, PA, USA}

\maketitle
\begin{abstract}
Studying animal locomotion improves our understanding of motor control and aids in the treatment of motor impairment. Mice are a premier model of human disease and are the model system of choice for much of basic neuroscience. High frame rates (250 Hz) are needed to quantify the kinematics of these running rodents. Manual tracking, especially for multiple markers, becomes time-consuming and impossible. Therefore, an automated method is necessary. We propose a method to track the paws of the animal in the following manner: first, segmenting all the possible paws based on color; second, classifying the segmented objects using a support vector machine (SVM) and neural network (NN); third, classifying the objects using the kinematic features of the running animal, coupled with texture features from earlier frames; and finally, detecting and handling collisions to assure the correctness of labelled paws. The proposed method is validated in sixty 1,000 frame video sequences (4 seconds) captured by four cameras from five mice. The total sensitivity for tracking of the front and hind paw is 99.70\% using the SVM classifier and 99.76\% using the NN classifier. In addition, we show the feasibility of 3D reconstruction using the four camera system. 
\end{abstract}
\begin{keywords}
Biomechanics, Image Processing, Rodent Tracking, Artificial Intelligence
\end{keywords}

\section{Introduction}
\label{intro}
Wireless capsule endoscopy (WCE) is a new device which is able to investigate the entire gastrointestinal (GI) tract without any pain. It captures more than 55000 frames during an examination (a minimum of two frames per second and a recording time of eight hours). Physicians need to spend a long time to review these frames. Therefore, it would be appropriate to reduce the review time by an automatic method. To date, various methods have been proposed to recognize bleeding or ulceration regions in the GI tract, but limited studies have been executed to recognize abnormal (diseases) and tumor tissues \cite{alizadeh2014segmentation}, \cite{billahgastrointestinal}. 

To detect the bleeding region in the WCE frames, a method was presented based on color similarity features \cite{guobing2011novel}. The pixels classified as bleeding pixels were used as seeds in a region-growing algorithm to find the entire bleeding regions. In another study, color features were extracted from the RGB and HSI color spaces and pixels classified using a probabilistic neural network \cite{pan2011bleeding}. Baopu and Meng presented a method based on chrominance moment as a color feature and uniform local binary patterns (LBP) to detect bleeding regions in a WCE frame \cite{li2009computer}. A three-layer perceptron neural network was used to detect bleeding pixel. Three classifiers were utilized to evaluate the method: support vector machine (SVM); linear discriminant analysis; and K-nearest neighbors (KNN).

To detect ulcer in the WCE frames, a method was proposed using Gabor filter, color and texture features, following by a neural network to classify the frames \cite{karargyris2009identification}. They also developed another method to detect ulcer and polyp frames in the WCE videos \cite{karargyris2011detection}. Their method was extracted features using log Gabor filters, color features, and Susan edge detector (a geometry feature). Then, the frames were classified using an SVM. 

A method was introduced to detect the frames that contained Crohn's disease by the edge histogram descriptor in four angles, color features based on the LUV color space, texture features extracted by a Gabor filter bank \cite{kumar2012assessment}. An SVM was employed to classify lesion tissues in a frame using the extracted features. Another method was presented using uniform LBP and discrete wavelet transform to detect tumor frames following by an SVM classifier \cite{li2012tumor}. Li et. al. presented another method based on multi-scale LBP and multiple classifiers (KNN, multi-layer perceptron (MLP) neural network, and SVM) \cite{li2011computer}. We proposed a method to segment abnormal regions in a frame based on the intensity value \cite{maghsoudi2012segmentation}. That method was sensitive to illumination and did not work well to detect Crohn's disease. 

In one of the recent studies \cite{szczypinski2014texture}, a method was proposed to detect erosion, ulcer, and bleeding regions. MaZda software was used to extract texture features from seven color spaces, and then, an SVM classified the frames. We presented methods to segment bubbled regions \cite{maghsoudi2014detection} and to distinguish different organs \cite{maghsoudi2012automatic} in the WCE frames which they might be helpful for future studies.

Overall, we may conclude that color features were most effective features for detection of bleeding regions while these features were not as effective for abnormal regions detection. None of the above studies used geometry, color, and texture features together for detection of abnormalities (expect ulceration and bleeding) in the WCE frames; we employed different types of features to improve the detection rate. Moreover, each method was devised to detect a specific type of disease in the GI tract. To overcome the above limitations, we propose two methods using the different texture, geometry, and color features to detect tumor frames (frame based study) and abnormal regions (pixel-based study). They have four advantages relative to previous methods: first, different features are extracted to assure the highest possible detection rate; second, three important classes of abnormalities are detected (tumor, bleeding, and a group of five other diseases); third, they improve upon the results of previous studies; fourth, to find the best features for this kind of detection of the three classes.

\section{Definitions of Diseases and Features}
\label{sec:1}
\subsection{Bleeding, Tumor, and Other Abnormalities}
\label{sec:2}
Our study is limited to three major abnormalities: tumor (especially sub-mucosal tumor), blood and bleeding, and other abnormal regions (lymphangiectasia, lymphoid hyperplasia, xanthoma, stenosis, and Crohn's disease) \cite{schmassmann2005handbook}. The etiology of the diseases is not clear; however, the early diagnosis can affect the treatment. Therefore, the development of techniques for detecting these lesions is important and reasonable.

\subsection{Features}
\label{sec:3}
\subsubsection{Gabor Filter}
The Gabor filter (G function) is same as a sinusoidal plane of frequency and orientation modulated by a Gaussian envelope before convolving with images. This filter has good localization properties in both spatial and frequency domains and has been used for texture segmentation \cite{jain1991unsupervised}, \cite{penjweini2017investigating}. The impulse response of the 2D Gabor filter is
\begin{equation}
G = \dfrac{e^{-(\dfrac{X^{2}}{2\sigma_{x}^{2}}+\dfrac{Y^{2}}{2\sigma_{y}^{2}})}}{2\pi\sigma_{x}\sigma_{y}} \times e^{i(2\pi fX+\phi})
\end{equation}
where
\begin{equation}
S(x)=\left\{
\begin{array}{   l   l    }
X = x {\times} cos(\theta)+y \times sin(\theta)\\
Y = -x \times sin(\theta)+y \times cos(\theta) \\
\end{array}
\right.
\end{equation}
and $\theta$ is the rotation angel of the impulse response, $x$ and $y$ are the coordinates, $\sigma_{x}$ and $\sigma_{y}$ are the standard deviations of the Gaussian envelope in the $x$ and $y$ directions, respectively, $f$ and $\theta$ are the frequency and phase of the sinusoidal, respectively. 

\subsubsection{Local Binary Patterns (LBP)}
\begin{figure}[t]
    \centering
        \includegraphics[width=0.25\textwidth]{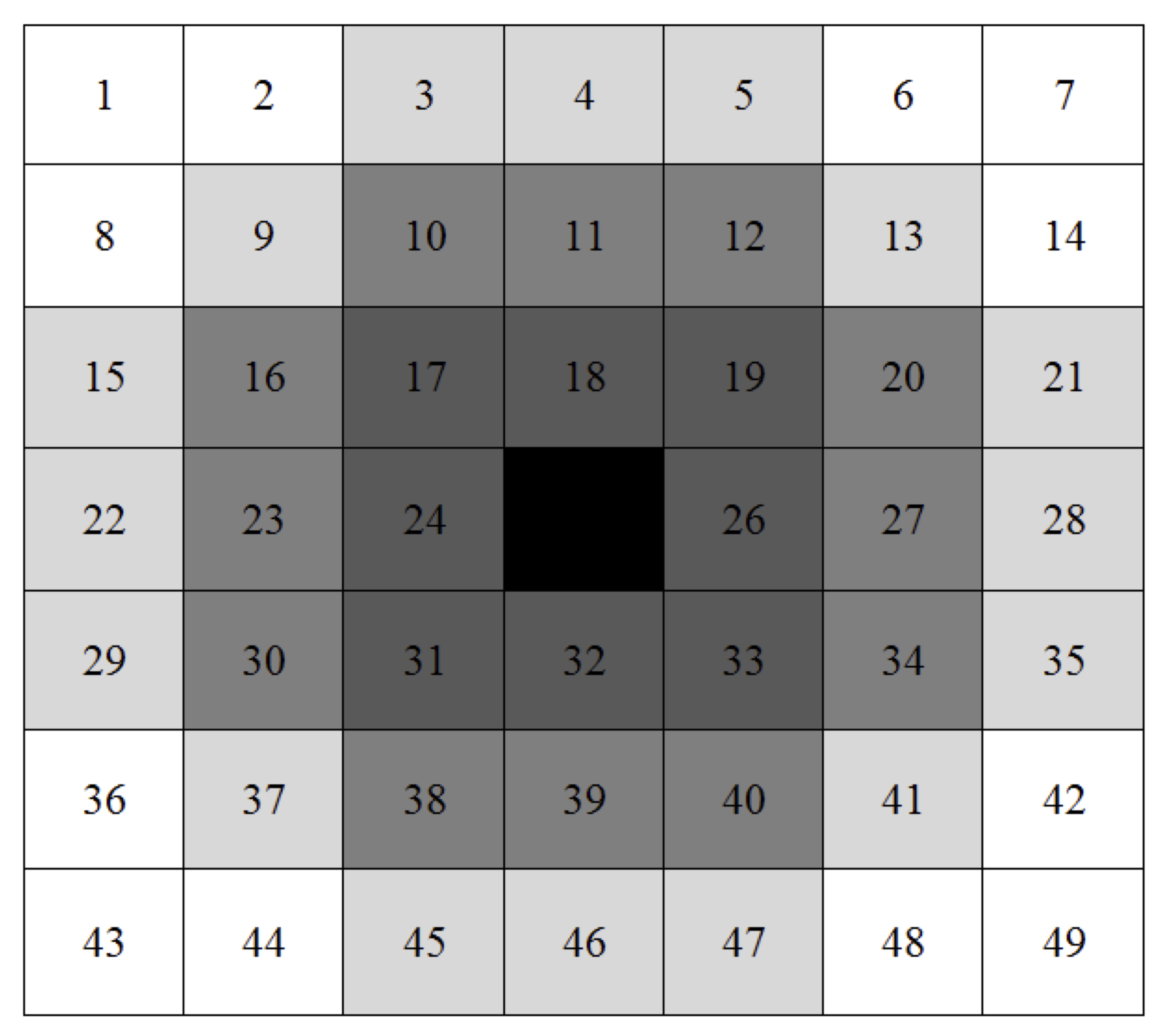}
    \caption{Image (a) shows LBP feature calculation for a sample block and Image (b) demonstrates correlation between radius and neighbours.}
    \label{fig:1}
\end{figure}

Local binary pattern has been widely used to extract texture feature \cite{rema2013segmentation}, \cite{li2009small}, \cite{maghsoudi2016detection}, \cite{ojala2002multiresolution}, \cite{yu2014improved}. The correlation between the number of neighbors and radius is described below and this correlation illustrated in Fig. \ref{fig:1} 
\begin{equation}
\left\{
\begin{array}{   l   l    }
$R = 1,2,3$\\
$Number of neighbors = 8,12,16$
\end{array}
\right.
\end{equation}

Then, two features are calculated based on LBP: LBP1 and LBP2. By rotating the LBP start point in a block to find all possible LBP values for a center, the minimum LBP is invariant to rotation. To extract LBP1 from rotation invariant LBP, three histograms are devised. LBP1 is counted based on the following ranges while selecting each range depends on the radius:
\begin{equation}
\left\{
\begin{array}{   l   l    }
$H1(i) =[1, 3, 7, 15, 31, 63, 127, 255, 511, 1023, 2047, $\\
$ \ \ \ \ \ \ \ \ \ \ 4095, 8191, 16383, 32768, 65535]$\\
$H2(j) =[1, 3, 7, 15, 31, 63, 127, 255, 511, 1023, 2047,  $\\
$ \ \ \ \ \ \ \ \ \ \ 4095]$\\
$H3(k) =[1, 3, 7, 15, 31, 63, 127, 255]$
\end{array}
\right.
\end{equation}
where $1 \leq  i \leq 16$, $1 \leq j \leq 12$, and $1\leq k \leq 8$. By counting repetition number of LBP1 values placed between two consecutive members of a histogram, 15 features are extracted for (neighbors and radius) = (16 and 3), 11 features for (neighbors and radius) = (12 and 2), 7 features for (neighbors and radius) = (8 and 1), and 4 features by calculating the percentage of LBP1 placed between H1(15) and H1(16), H1(14) and H1(15), H2(11) and H2(12), and H3(7) and H3(8). Therefore, 37 features are extracted using the LBP1 histograms. 

To extract LBP2 from LBP (not rotation invariant LBP), another histogram is used to find which neighbor is more important (which one repeats further than others). To calculate this new feature, neighbors are labeled by the following series that are demonstrated in  Fig. \ref{fig:1}: 
\begin{equation}
\left\{
\begin{array}{   l   l    }
$C1(i) = [3, 4, 5, 13, 21, 28, 35, 41, 47, 46, 45, 37, 29, $\\
$ \ \ \ \ \ \ \ \ \ \ 22, 15, 9] $\\
$C2(j) = [10, 11, 12, 20, 27, 34, 40, 39, 38, 30, 23, 16] $\\
$C3(k) = [17, 18, 19, 26, 33, 32, 31, 24] $
\end{array}
\right.
\end{equation}
and the order of neighbors is defined using the following series:
\begin{equation}
\left\{
\begin{array}{   l   l    }
$O1(j) =  [1, 2, 3, 4, 5, 6, 7, 8, 9, 10, 11, 12, 13, 14,  $\\
$ \ \ \ \ \ \ \ \ \ \ 15, 16]  $\\
$O2(j) = [1, 2, 3, 4, 5, 6, 7, 8, 9, 10, 11, 12] $\\
$O3(k) = [1, 2, 3, 4, 5, 6, 7, 8]$
\end{array}
\right.
\end{equation}
Therefore, 36 features are extracted using the LBP2 histograms.

\subsubsection{Law's Features}
Law's texture features have been used widely \cite{laws1980textured}. These features were developed by Kenneth Ivan Law at the University of Southern California. 21 Law's masks for 7 samples and 15 Law's masks for 5 samples were applied to the frames.

\subsubsection{GLCM Features}
Fourteen Haralick features \cite{sahu2015classification} are extracted from a gray level co-occurrence matrix (GLCM): contrast; correlation; entropy; energy; difference variance; difference entropy; information measure of correlation 1; information measure of correlation 2; inverse difference; sum average; sum variance; sum of squares; sum entropy; and maximum correlation coefficient. 

Additional features that are extracted from GLCM are \cite{soh1999texture}, \cite{clausi2002analysis}: auto correlation; cluster prominence; cluster shade; dissimilarity; homogeneity; maximum probability; inverse difference normalized; and inverse difference moment normalized. Totally, 22 features are extracted from GLCM. In this study, if the directions and distances of GLCM are not mentioned, direction and distance are 0 and 1, respectively.

\subsubsection{Invariant Moments}
The following invariant moments are computed based on the information provided by both shape boundary and its interior regions \cite{chen1993improved}: 
\begin{equation}
M_{pq} = \Sigma_{x}\Sigma_{y}x^{p}y^{q}f(x,y)
\end{equation}
where $M_{pq}$ is the two-dimensional moment of the function $f(x,y)$. The order of the moment is $(p + q)$ where $p$ and $q$ are both natural numbers. The following central moments are defined to generate features that are invariant to translation:
\begin{equation}
\overline{y} = \dfrac{M_{01}}{M_{00}}, \ \overline{x} = \dfrac{M_{10}}{M_{00}}.
\end{equation}
In the discrete domain, these moments become:
\begin{equation}
\mu_{pq} = \Sigma_{x}\Sigma_{y}(x-\overline{x})^{p}(y-\overline{y})^{q}
\end{equation}
The moments are further normalized for the effects of scale change using:
\begin{equation}
\tau_{pq} = \dfrac{\mu_{pq}}{\mu_{pq}^{\gamma}}
\end{equation}
\begin{figure}[t]
    \centering
        \includegraphics[width=0.5\textwidth]{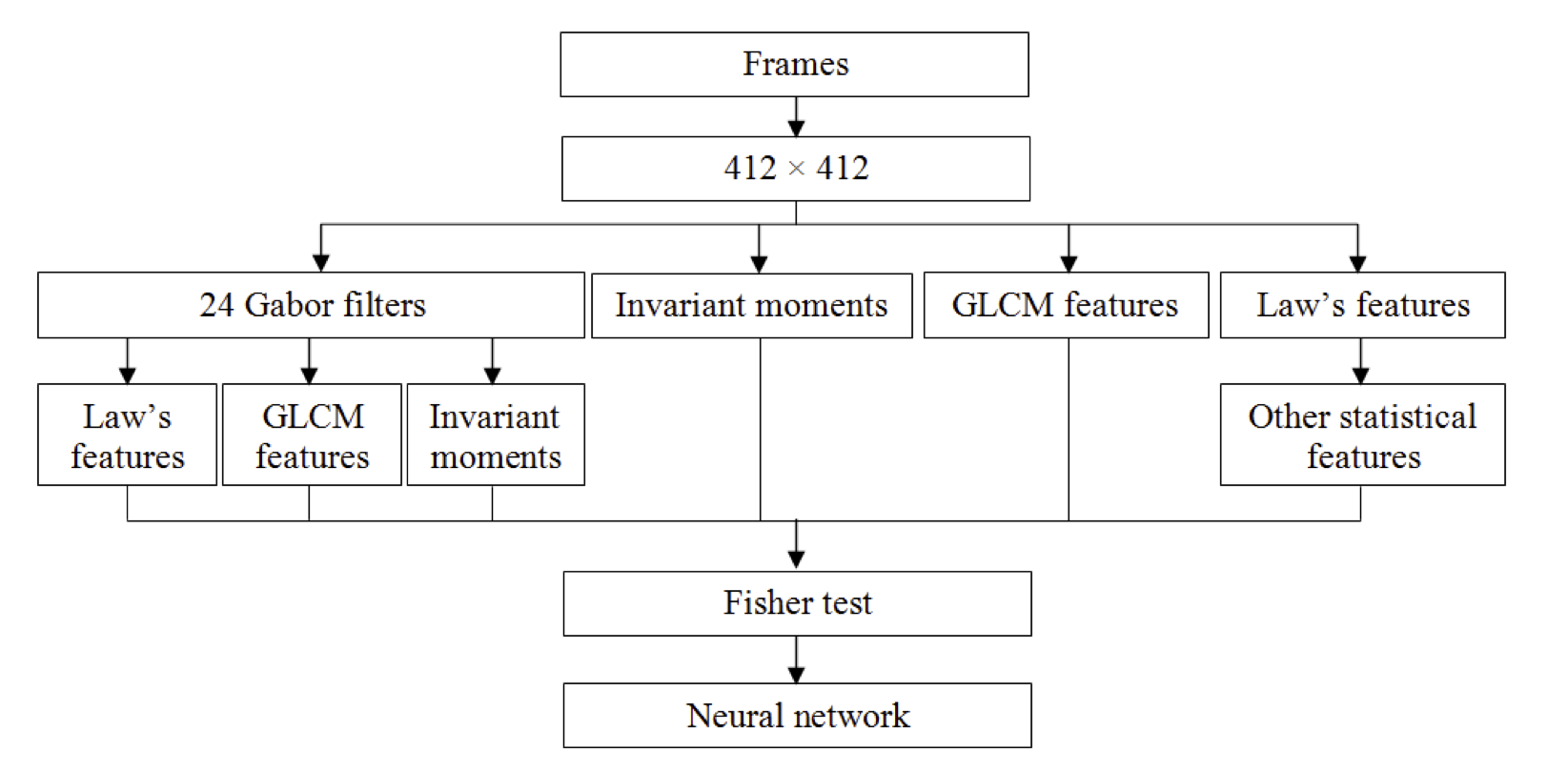}
    \caption{The first algorithm to detect tumor frame in a video.}
    \label{fig:2}
\end{figure}
where $\gamma = (p + q) / 2 + 1$. From the normalized central moments, the following invariant moments (to scale, translation, and rotation) can be calculated \cite{hu1962visual}:
\begin{equation}
\begin{array}{   l   l    }
\phi_{1}=\tau_{20}+\tau_{02} \\
\phi_{2}=(\tau_{20}-\tau_{02})^{2}+4\tau_{11}^{2}  \\
\phi_{3}=(\tau_{30}-3\tau_{12})^{2}+(\tau_{03}-3\tau_{21})^{2}\\
\phi_{4}=(\tau_{30}+\tau_{12})^{2}+(\tau_{03}+\tau_{21})^{2} \\
\phi_{5}=(3\tau_{30}-3\tau_{12})(\tau_{30}+\tau_{12})[(\tau_{30}+\tau_{12})^{2}-3(\tau_{03}+\tau_{21})^{2}]\\
\ \ \ \ \ \ + (3\tau_{21}-3\tau_{03})(\tau_{21}+\tau_{03})[3(\tau_{30}+\tau_{12})^{2}-(\tau_{21}+\tau_{03})^{2}] \\
\phi_{6}=(\tau_{20}-\tau_{02})[(\tau_{30}+\tau_{12})^{2}-(\tau_{21}+\tau_{03})^{2}]\\
\ \ \ \ \ \ + 4\tau_{11}(\tau_{30}+\tau_{12})(\tau_{21}+\tau_{03}) \\
\phi_{7}=(3\tau_{21}-\tau_{03})(\tau_{30}+\tau_{12})[(\tau_{30}+\tau_{12})^{2}-3(\tau_{03}+\tau_{21})^{2}]\\
\ \ \ \ \ \ + (3\tau_{21}-3\tau_{03})(\tau_{21}+\tau_{03})[3(\tau_{30}+\tau_{12})^{2}-(\tau_{21}+\tau_{03})^{2}] 
\end{array}
\end{equation}

\section{Proposed Methods}
\subsection{Detection of Frames Showing Tumor}
To find tumor frames in a video, the following steps were applied and also illustrated in Fig. \ref{fig:2}: 
\begin{enumerate}[(A)]
\item Frame size was $512 \times 512$ in our data set. As discussed, geometry features were useful in detecting frames showing tumor while the borders interfered with this task. Therefore, a central region of $412 \times 412$ pixels was extracted from each frame for farther processing. 
\end{enumerate}
\begin{enumerate}[(B)]
\item The Gabor filter was applied using the following parameters:
\begin{equation}
\begin{array}{   l   l    }
f = [0.5,1.5]; \theta = [0, 45, 90, 135]; \\
$x and y ranges$ = [10,20,40]; \phi = [0]; \\
\sigma_{x} =[.5]; \sigma_{y} =[.5]
\end{array}
\end{equation}
This process generated 24 images. Then, each set of 4 images generated using each of the above $\theta$ values were averaged to generate six additional images. 
\end{enumerate}
\begin{enumerate}[(C)]
\item For each of the images generated above, 22 GLCM features, 7 invariant moments, and 4 additional statistical features (mean, variance, skewness, and kurtosis) were extracted. The total number of features was 990 (30 images generated using the Gabor filters $\times$ 33 features = 990 features).
\end{enumerate}
\begin{enumerate}[(D)]
\item In addition to the above, 75 Law's features were extracted as follows: mean, variance, skewness, kurtosis, and entropy of 15 images, generated by convolving the image with 5-sample Law's masks (15 masks). 88 features were extracted from GLCM calculated for four different angles and 7 invariant moments were calculated. These features were extracted from the gray scale version of each original frame. Totally, 1160 features (990+75+88+7) were extracted. 
\end{enumerate}
\begin{enumerate}[(E)]
\item The features were normalized between zero and one because different features extraction methods were used. Thirty most discriminant features for detection of tumor frames were found using the Fisher Test \cite{fisher1915mathematical}. A multi-layer Perceptron (MLP) neural network \cite{mohapatra2012lymphocyte} was used to classify the WCE frames.
\end{enumerate}
\begin{figure}[t]
    \centering
        \includegraphics[width=0.5\textwidth]{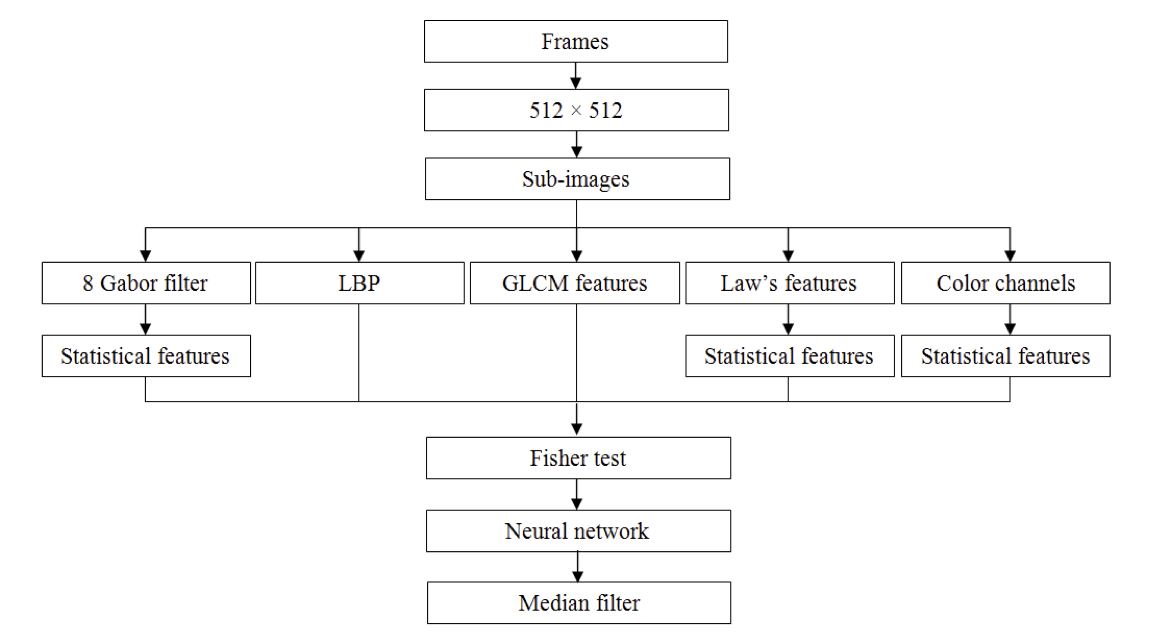}
    \caption{The second algorithm to detect abnormal regions in a frame.}
    \label{fig:3}
\end{figure}

\subsection{Detection of Pixels Showing Bleeding, Tumor, and Other Abnormalities}
This algorithm separates normal from abnormal tissues (bleeding, tumor, and other abnormalities). Fig. \ref{fig:3} shows the overview of this approach and the steps are described below: 
\begin{enumerate}[(A)]
\item Then, each frame was divided into 256 $32 \times 32$ sub-images. 
\end{enumerate}
\begin{enumerate}[(B)]
\item A sub-image size was $32 \times 32$ and an LBP block was $7 \times 7$; therefore, 26 rows and columns were possible for being the center of a block. Therefore, LBP was calculated 676 times for a sub-image (26 rows $\times$ 26 columns). 74 features were extracted using LBP1 from the grayscale and green channel for each sub-image. Moreover, 36 features were extracted using LBP2 from the grayscale version of sub-images. 
\end{enumerate}
\begin{enumerate}[(C)]
\item The mean and GLCM features were extracted from the grayscale sub-images for four angles (0, 45, 90, and 135). Therefore, the extracted features increased to 110 + (22+1) $\times$ 4=202. 
\end{enumerate}
\begin{enumerate}[(D)]
\item As discussed in Section 2.2.3, 21 two-dimensional masks (for 7 samples) were generated using the one-dimensional Law's kernels. The mean, variance, skewness, kurtosis, and entropy were extracted from the convolved sub-image with these masks. 21 $\times$ 5 = 105 features were extracted in this step.
\end{enumerate}
\begin{enumerate}[(E)]
\item Eight Gabor filters were generated by two frequencies and four angles to apply on a sub-image. Moreover, each set of 4 images generated using the Gabor filters, same as the first method, were averaged to generate two additional images. Then, the mean, variance, skewness, kurtosis, and entropy were extracted from the generated images. Therefore, features extracted in this step were 50 (5 $\times$ (8+2)).
\end{enumerate}
\begin{enumerate}[(F)]
\item HSV color space has been a popular color space to detect objects using color information for different applications \cite{maghsoudi2012automatic},\cite{junzhou2011contourlet},\cite{maghsoudi2016tracker},\cite{pan2011bleeding}. The mean, variance, skewness, and kurtosis were extracted from five color channels (red, green, blue, hue, and saturation) and the grayscale sub-image. Totally, 202 (LBP) + 105 (Law) + 50 (Gabor) + 24 (color channels + grayscale) = 381 features were extracted.
\end{enumerate}
\begin{enumerate}[(G)]
\item The features ranges extracted above were normalized between zero and one because different features extraction methods were used. Then, the number of features was reduced to 30 using the Fisher test \cite{fisher1915mathematical}; the selected features were the most discriminant features for distinguishing normal and abnormal regions in a frame. 
\end{enumerate}
\begin{enumerate}[(H)]
\item Three networks with three hidden layers were employed to classify normal and abnormal regions \cite{leondes1998neural}. The tumor neural network, other abnormalities neural network, and bleeding neural network performances were respectively 0.1028, 0.0548, and 0.0247.
\end{enumerate}
\begin{figure}[t]
    \centering
        \includegraphics[width=0.45\textwidth]{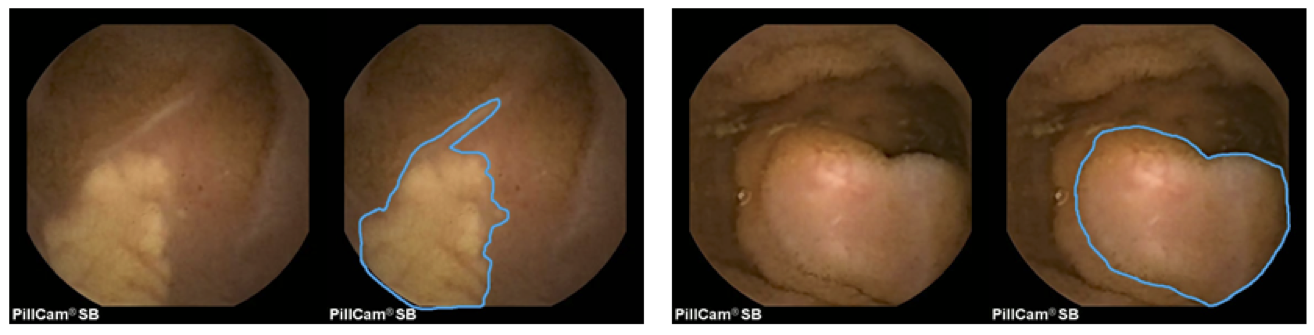}
    \caption{Normal and abnormal regions determined by a specialist in two sample frames.}
    \label{fig:4}
\end{figure}
\section{Results}
In this study, 233 frames taken from 59 patients were used. The videos were captured by the M2A capsule endoscopy device manufactured by the Given Imaging Company and provided by the Shariati Hospital, Tehran, Iran. From these videos, 43 tumor frames from 11 patients, 44 normal frames from 12 patients, 33 bleeding frames from 9 patients, and 113 other abnormalities frames were selected from 29 patients. Lymphangiectasia, stenosis, lymphoid hyperplasia, xanthoma, and Crohn's disease had respectively 18 frames from 5 patients, 31 frames from 6 patients, 19 frames from 4 patients, 17 frames from 6 patients, and 28 frames from 8 patients. 

The specialist, Dr. Soleimani, supervised the process to segment the normal and abnormal regions in the frames. Fig. \ref{fig:4} shows normal and abnormal regions separated in two frames. Then, random sampling and cross-validation over the patient's frames were used to divide the frames into the training (approximately 75\%) and testing sets (approximately 25\%). 
\begin{table}[t]
  \centering
  \caption{Performance of the first method in detecting tumor frames from the WCE videos. }
    \begin{tabular}{ccc}
    \toprule
          & Tumor (TEST) & Normal (TEST) \\
    \midrule
    Correct decisions & 12    & 41 \\
    \midrule
    Wrong decisions & 0     & 2 \\
    \bottomrule
    \end{tabular}
  \label{tab:1}
\end{table}
\begin{table}[b]
  \centering
  \caption{Sensitivity and specificity of the first method for different feature sets found by the Fisher test.}
    \begin{tabular}{c|c|c}
    \toprule
    Number of features & Sensitivity & Specificity \\
    \midrule
    10    & 0.6667 & 0.7241 \\
    \midrule
    20    & 0.8333 & 0.8276 \\
    \midrule
    25    & 0.8889 & 0.8621 \\
    \midrule
    30    & 1     & 0.951 \\
    \midrule
    35    & 1     & 0.951 \\
    \midrule
    40    & 1     & 0.928 \\
    \bottomrule
    \end{tabular}
  \label{tab:2}
\end{table}

\subsection{Detection of Frames Showing Tumor}
In this case, 43 tumor frames (31 frames for training) and 161 non-tumor frames (120 frames for training) were selected. Non-tumor frames were 44 normal and 117 other abnormalities frames. The aim was to find frames that contain tumor regions in the testing samples (totally 12 tumor frames from 4 patients+ 41 non-tumor frames from 13 patients). Bleeding frames were not counted here because some bleeding frames were not obvious (because of the bleeding regions). If bleeding frames were mixed by non-tumor frames, more features should have been extracted to distinguish them from tumor frames, especially color features; moreover, some tumor frames contain bleeding regions (kind of confusion in classification). 

The usual method to determine the Gabor filter parameters (Eq. 12) is to apply a large bank of Gabor filter covering all possible frequencies, $\theta$, $\phi$, and x and y ranges. For frequency, the range was between 0.5 and 2.5 and we considered two samples. The x and y ranges were selected in a way that the output image seemed meaningful by three steps \cite{jain1991unsupervised}. Four spatial angles were considered to examine different orientations. 

\begin{figure}[t]
    \centering
        \includegraphics[width=0.45\textwidth]{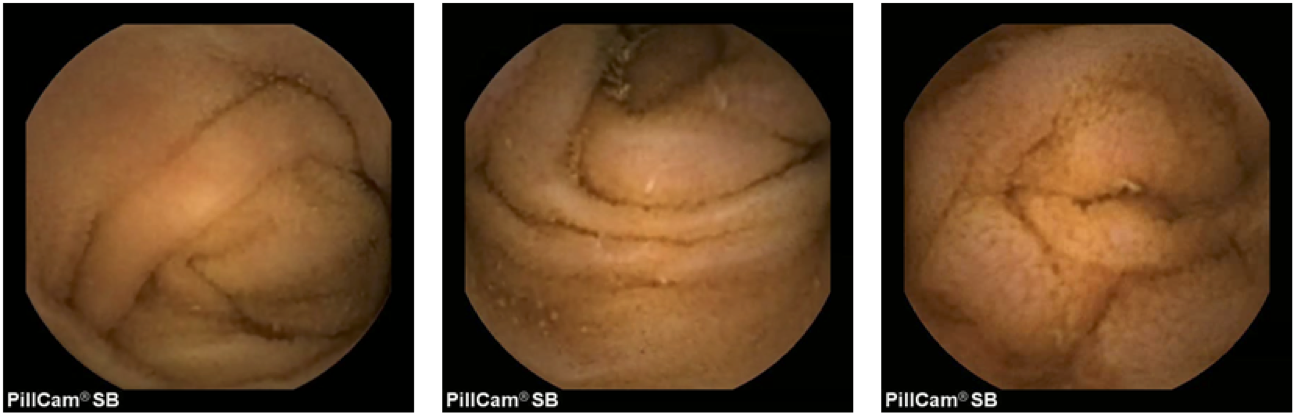}
    \caption{Left and center images show two normal frames from the testing samples that are classified incorrectly and the right image shows a normal frame that is correctly classified.}
    \label{fig:6}
\end{figure}

The shape was one of the main features for detection of tumor frames. In addition to geometry features, texture and color features of a region were important. Therefore, the Gabor filter was applied and features were extracted from the generated images. The results are illustrated in Table 1. Two normal frames indicated as tumor frame are demonstrated in Fig. \ref{fig:6}. 

The Fisher test was used to find thirty most discriminating features (out of 1160 features) from the frame. Table 2 shows the effect of the number of features on the sensitivity and specificity of the first method, indicating that 30 features are optimal (regarding the feature pace = 5). 

\subsection{Detection of Pixels showing Bleeding, Tumor, and Other Abnormalities}
Here, 12 tumor frames, 10 completely normal frames, 7 bleeding frames, and 31 other abnormalities frames generated the dataset for evaluation of the second method. However, for example, the actual dataset for detection of tumor pixels was (12 tumor frames + 10 normal frames) $\times$ 512 $\times$ 512 pixels = 2,621,452 pixels. 

After using the neural networks, each frame was smoothed by a median filter. The median filter was applied on a frame with 25-pixel samples (window size). The methods were estimated by measuring sensitivity, specificity, accuracy, and precision \cite{altman1994diagnostic}. 

\begin{figure*}[t]
    \centering
        \includegraphics[width=0.85\textwidth]{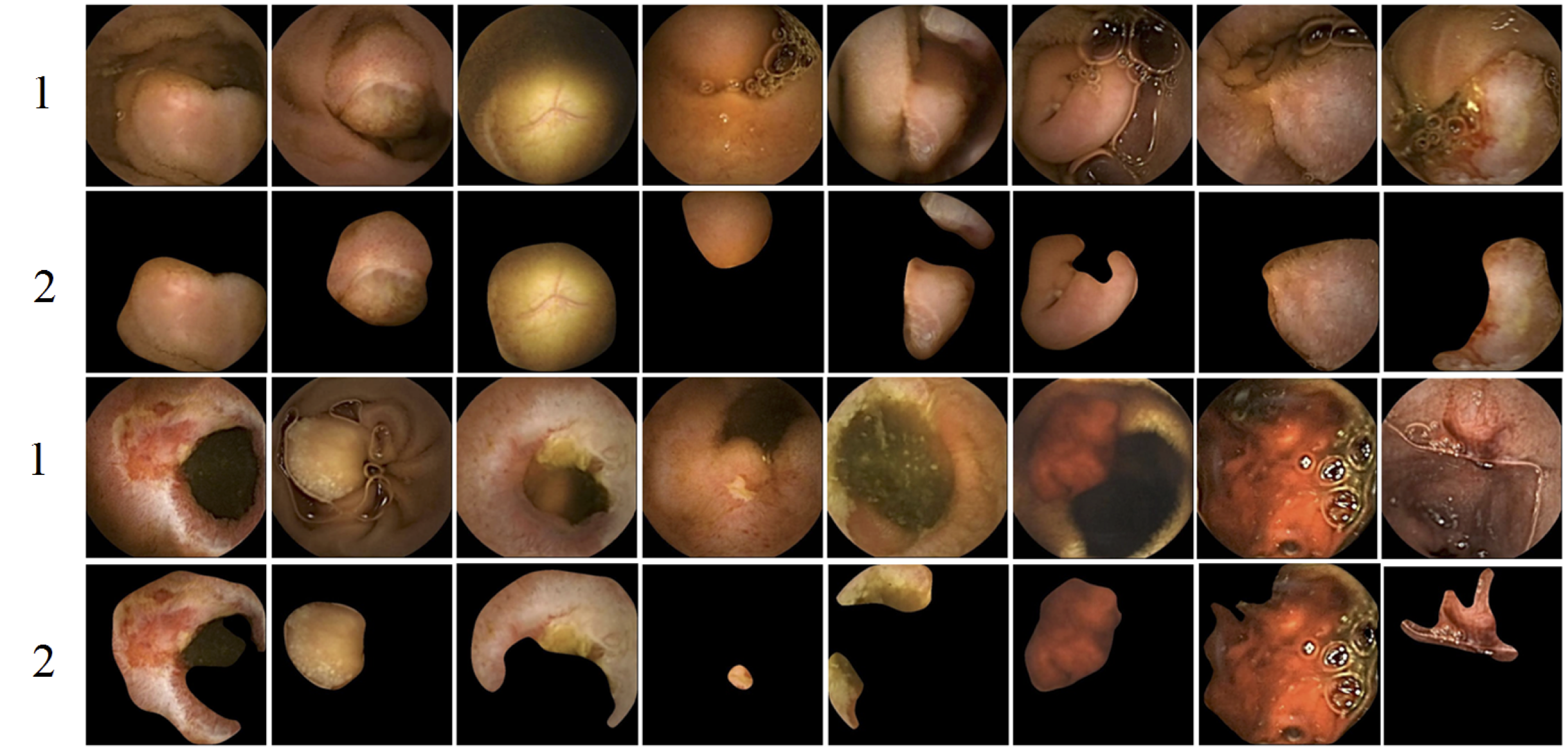}
    \caption{The two rows labeled as 1 show the original frames and the two rows labeled as 2 show the segmented regions.}
    \label{fig:7}
\end{figure*}

\begin{figure}[b]
    \centering
        \includegraphics[width=0.45\textwidth]{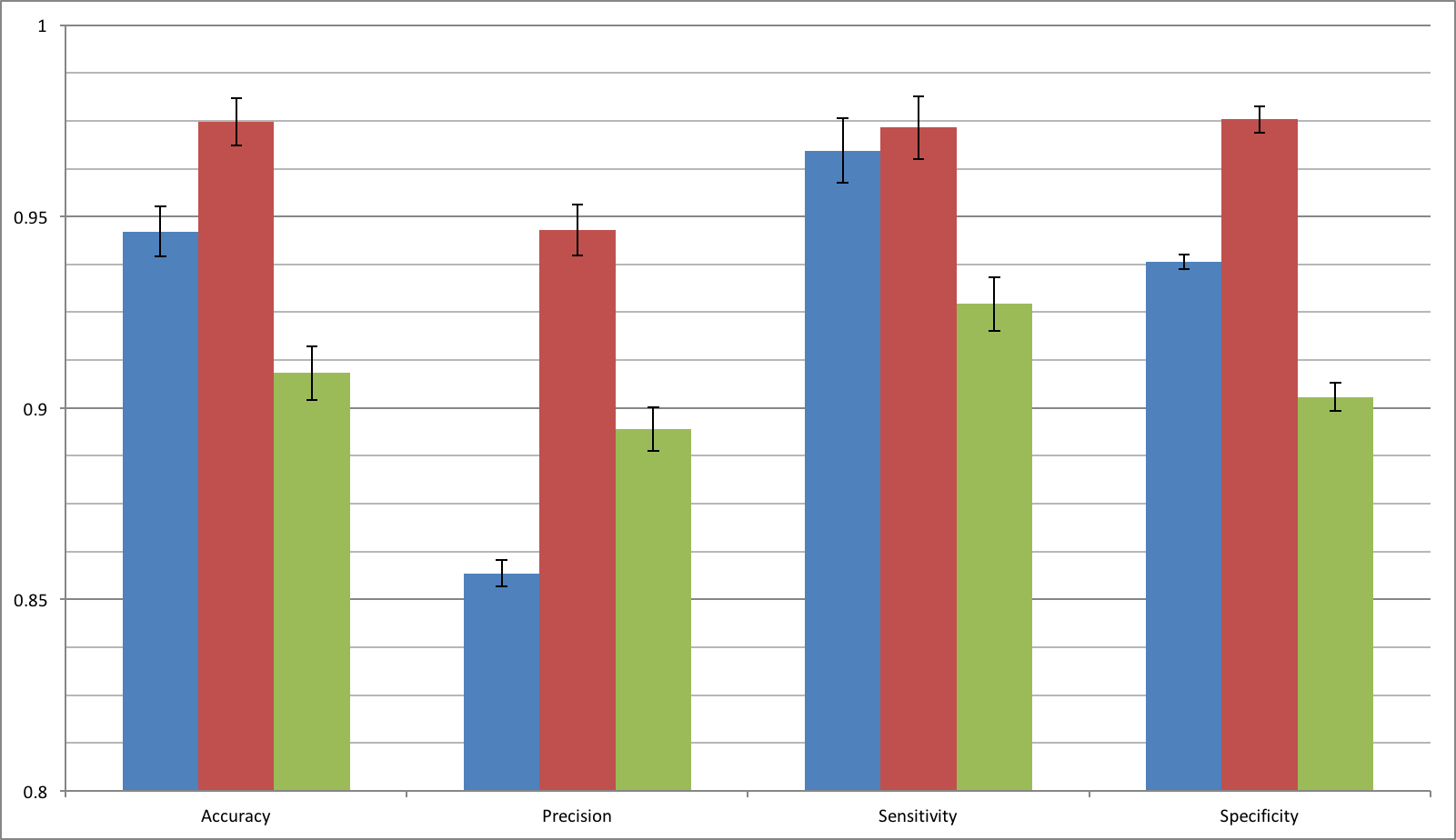}
    \caption{Accuracy, precision, sensitivity, and specificity of segmenting tumor, bleeding, and other abnormalities are shown in blue, red, and green bars, respectively. The error bars show the standard deviation.}
    \label{fig:8}
\end{figure}

\begin{table}[b]
  \centering
  \caption{Performance of the second method in detecting abnormal regions from the WCE frames.}
    \begin{tabular}{c|c|c|c}
    \toprule
          & Tumor & Bleeding & Abnormalities \\
    \midrule
    Accuracy & 0.9092 & 0.9747 & 0.9461 \\
    \midrule
    Precision & 0.8945 & 0.9465 & 0.8568 \\
    \midrule
    Sensitivity & 0.9273 & 0.9733 & 0.9671 \\
    \midrule
    Specificity & 0.9029 & 0.9754 & 0.9381 \\
    \bottomrule
    \end{tabular}%
  \label{tab:3}%
\end{table}%
\begin{table}[b]
  \centering
  \caption{Sensitivity and specificity of the second method (only for detecting tumor regions) for different feature sets found by the Fisher test.}
    \begin{tabular}{c|c|c}
    \toprule
    Number of features & Sensitivity & Specificity \\
    \midrule
    20    & 0.8521 & 0.8633 \\
    \midrule
    25    & 0.9011     & 0.8821 \\
    \midrule
    30    & 0.9273     & 0.9029 \\
    \midrule
    35    & 0.9251     & 0.9022 \\
    \midrule
    40    & 0.9263     & 0.9011 \\
    \bottomrule
    \end{tabular}%
  \label{tab:4}%
\end{table}%
Fig. \ref{fig:7} demonstrates how the second method distinguishes normal and abnormal regions in some sample frames and Table 3 shows these measures for the testing samples. In addition to Table 3, these results are illustrated as a chart in Fig. \ref{fig:8}. 

Same as the first method, the Fisher test was used to find 30 most discriminating features (out of 381 features) for each sub-image. Table 4 shows the effect of a number of features on the sensitivity and specificity of the second method (only for detecting tumor regions), indicating that 30 features were optimal; these 30 features were illustrated in Table 5. As illustrated, using 35 features reduced the performance because some features caused confusion in making a decision. Moreover, using 25 features reduced the performance because some discriminating features were absent. 

The methods were examined using Matlab 2016a on a MacBook pro 2.7 GHz Intel Core i5 with 8 GB 1867 MHz DDR3. The average required time to process a frame was $0.178\pm 0.036$ second (mean $\pm$ standard deviation) and this time was $0.762\pm 0.073$ second.

\section{Conclusion}
Two methods were proposed: the first method can find tumor frames in a video (or a dataset) and the second one can recognize normal from bleeding regions, normal from tumor regions, and normal from a group of five abnormalities regions in a frame. Geometrical information was one of the main features to distinguish tumor and normal tissues in a frame. Therefore, invariant moments were used to extract shape features once from the created frame after using the Gabor filter bank and then, from the grayscale version of the original frame. Moreover, GLCM, statistical, and Law's features were used twice; first, to show an impact of texture feature for detection of tumor frames and secondly, to extract geometry features after applying the Gabor filters. This method indicated respectively 100\%, 96\%, 93\%, and 90\% as sensitivity, accuracy, specificity, and precision for detection of frames showing the tumor. Invariant moments helped us to achieve well results. 

\begin{table}[b]
  \centering
  \caption{The following features were the final ones for the training of each neural network. TEF, SFB, SFT, and SFD are total extracted features, selected features for bleeding, selected features for tomur, selected features for diseases, respectively.}
    \begin{tabular}{c|c|c|c|c}
    \toprule
          & LBP   & GLCM & Gabor & Color \\
    \midrule
    TEF & 110   & 92    & 50    & 24 \\
    \midrule
    SFB & 3     & 12    & 0     & 15 \\
    \midrule
    SFT & 12    & 6     & 7     & 5 \\
    \midrule
    SFD & 10    & 10    & 3     & 7 \\
    \bottomrule
    \end{tabular}
  \label{tab:5}
\end{table}
It is important to find frames containing abnormalities; however, it would be helpful to find abnormal regions in a frame. The second method was presented to distinguish normal and abnormal regions in a frame. In this case, texture features were more helpful; therefore, GLCM, statistical, LBP (LBP1 and LBP2), and Law's features were used. The results of this method are illustrated in Table 3. 

To summarize the significance of our study, as Table \ref{tab:6} shows, the frame based detection method for tumor achieved to a significantly higher sensitivity compared with the tumor and polyp detection methods. Although, our method needs to be evaluated for more tumor frames as our testing dataset was limited to 12 frames. We will attempt to evaluate the method by gathering more frames in future works. On the other hand, we had enough data set to evaluate our pixel-based method. The pixel-based method to detect bleeding achieved a slightly higher specificity compared to the previous studies. In addition, the pixel-based method to detect tumor and abnormalities showed an improvement in the reported results by the previous studies. 

\begin{table*}[tp]
\caption{Comparison between the available methods and our proposed methods. The first column shows the method type which can be classification after segmentation or only segmentation. FB, PB, and NR are abbreviations for frame based method, pixel based method, and results which have not reported.} 
\label{tab:6}
\begin{center}   
\scalebox{0.8}{
\begin{tabular}{| c | c | c | c | c | c | c | c |}
    \hline
    Method & Mode & Aim & Sensitivity & Specificity & Accuracy & Precision\\ \hline
       Classification{\cite{pan2011bleeding}} & FB & Bleeding & 0.97 & 0.91 & NR & NR \\ \hline
       Classification {\cite{li2011computer}} & FB & Tumor & 0.85 & 0.80 & 0.82 & NR \\ \hline
    Classification{\cite{li2011computer}} & FB & Tumor & 0.87 & 0.84 & 0.86 & NR \\ \hline
    Classification{\cite{li2011computer}} & FB & Tumor & 0.88 & NR & 0.86 & 0.88 \\ \hline
    Classification{\cite{cui2010detection}} & FB & Lymphoid & 0.94 & NR & 0.55 & NR \\ \hline
    Classification{\cite{maghsoudi2016detection}} & FB & Polyps & 0.87 & 0.94 & 0.93 & NR \\ \hline
       Classification{\cite{karargyris2011detection}} & FB & Polyps & 1.00 & 0.68 & NR & NR \\ \hline
       Classification{\cite{karargyris2011detection}} & FB & Ulcer & 0.75 & 0.73 & NR & NR \\ \hline
       \textbf{Our Method} & \textbf{FB} & Tumor & 1.00 & 0.95 & 0.96 & 0.85 \\ \hline
       Classification{\cite{pan2011bleeding}} & PB & Bleeding & 0.92 & 0.88 & NR & NR \\ \hline
       Classification{\cite{silva2014toward}} & PB & Polyps & 0.91 & 0.95 & NR & NR \\ \hline
       Classification{\cite{karargyris2011detection}} & PB & Ulcer & 0.88 & 0.84 & NR & NR \\ \hline
       Classification{\cite{karargyris2011detection}} & PB & Polyps & 0.96 & 0.70 & NR & NR \\ \hline
       Classification{\cite{kumar2012assessment}} & PB & Crohn's & NR & 0.91 & 0.93 & 0.91 \\ \hline
    Segmentation{\cite{maghsoudi2012segmentation}} & PB & Crohn's & 0.89 & 0.65 & 0.75 & NR \\ \hline
    Segmentation{\cite{maghsoudi2012segmentation}} & PB & Lymphoid & 0.87 & 0.80 & 0.81 & NR \\ \hline
    \textbf{Our Method} & \textbf{PB} & Bleeding & 0.97 & 0.97 & 0.97 & 0.94  \\ \hline
    \textbf{Our Method} & \textbf{PB} & Diseases & 0.96 & 0.93 & 0.94 & 0.85 \\ \hline
    \textbf{Our Method} & \textbf{PB} & Tumor & 0.91 & 0.93 & 0.92 & 0.88 \\ \hline
\end{tabular}
}
\end{center}
\end{table*}
Deep learning has been used widely in image processing applications \cite{mansourdeep}. The main advantage of the presented work here compared to a convolutional neural network (CNN) is that this study showed the importance of a wide range of features (texture, geometry, and color features) to detect different types of abnormalities in the WCE frames while a CNN can be trained for a specific dataset and the internal features used by a CNN cannot be determined to extend the application. On the other hand, we needed to gather far more frames to train a CNN.  In addition, The method presented here can be used to classify the regions for deep learning applications as finding ROI in the frames is time consuming for physicians. In future work, we will try to collect more dataset to train and test a CNN and use some methods to eliminate the redundant image \cite{li2014online}, \cite{maghsoudi2014detection}. It might promise a method to segment abnormalities regions using a real-time process that may assist the WCE manufactures to add biopsy and drug delivery to the WCE. 

\section{acknowledgments}
We have been grateful for getting help from Dr. Hossein Asl Soleimani and Shariati Hospital for sharing the WCE frames with us.

\bibliographystyle{elsarticle-num}
\bibliography{report}

\end{document}